\documentclass{article} 
\usepackage{iclr2021_conference,times}

\usepackage{amsmath,amsfonts,bm}

\def\eqref#1{equation~\ref{#1}}

\def\1{\bm{1}}

\def\vc{{\bm{c}}}

\def\vg{{\bm{g}}}
\def\vh{{\bm{h}}}

\def\vx{{\bm{x}}}
\def\vy{{\bm{y}}}

\def\mR{{\bm{R}}}

\def\mW{{\bm{W}}}

\DeclareMathAlphabet{\mathsfit}{\encodingdefault}{\sfdefault}{m}{sl}
\SetMathAlphabet{\mathsfit}{bold}{\encodingdefault}{\sfdefault}{bx}{n}

\newcommand{\DCT}{\mathrm{DCT}}
\newcommand{\TopLeft}{\mathrm{TopLeft}}

\newcommand{\LSTM}{\mathrm{LSTM}}

\usepackage{wrapfig}

\usepackage{hyperref}
\usepackage{url}
\usepackage{booktabs}       
\usepackage{graphicx}
 \usepackage{subcaption, floatrow}
\usepackage{multirow}

\title{Training and Generating Neural Networks in Compressed Weight Space}

\author{Kazuki Irie, J\"urgen Schmidhuber\\
The Swiss AI Lab IDSIA, USI \& SUPSI \\
Lugano, Switzerland \\
\texttt{\{kazuki, juergen\}@idsia.ch}
}

\iclrfinalcopy 
\begin{document}

\maketitle

\begin{abstract}
The inputs and/or outputs of some neural nets are
weight matrices of other neural nets.
Indirect encodings or end-to-end compression of weight matrices
could help to scale such approaches.
Our goal is to open a discussion on this topic, starting with
recurrent neural networks for character-level language modelling whose weight matrices are encoded by the discrete cosine transform.
Our fast weight version thereof uses a recurrent neural network to parameterise
the compressed weights.
We present experimental results on the enwik8 dataset.
\end{abstract}

\section{Introduction}
Many modern neural nets (NNs) are big. 
Their number of trainable parameters may easily exceed a few hundred million.
This is a bottleneck for end-to-end differentiable systems that manipulate some NN's weights as input or output variables of another NN, i.e.,
fast weight programmers \citep{Schmidhuber:91fastweights, schmidhuber1992learning}.
This concept has seen a revival
under various names such as ``hyper-networks" \citep{ha2017hypernetworks},
and also in the context of Transformers with linear attention \citep{katharopoulos2020transformers, schlag2021linear}.
The former were presented as NNs that produce a weight matrix of another NN, but practical implementations 
only generate a vector that scales the rows of a weight matrix, instead of generating a de novo weight matrix.
In the original work  \citep{Schmidhuber:91fastweights, schmidhuber1992learning} and related fast weight variants 
\citep{schmidhuber1993reducing, ba2016using, schlag2017gated},
the matrix size bottleneck is addressed through outer products
of two NN-invented vectors of activations
which generate rank-one weight matrices at each time step.

For small NNs,
the whole weight matrix can be directly parametrised as the output of another NN (e.g.~\citet{Schmidhuber:91fastweights,sitzmann20}),
which is a natural way of generating context-dependent weight matrices.
If we had a differentiable compression method for weight matrices,
we could scale up such direct approaches.
Compact weight matrix representations could also be useful in  scenarios
which require a NN to read the weight matrix of another NN or itself 
\citep{Schmidhuber:92selfref, Schmidhuber:93selfreficann, Schmidhuber:93selfrefann, Schmidhuber:93selfrefieee, unterthiner2020predicting, harb2020policy, faccio2020parameter}.

In this work we revisit indirect encodings 
\citep{koutnik2010evolving, koutnik2010searching} based on the Discrete Cosine Transform (DCT) \citep{ahmed1974discrete}.
DCT offers a simple and differentiable encoding of NN weight matrices.
The model can be trained from scratch and end-to-end (including the DCT transformations).
While NNs with DCT-encoded weight matrices
have been successfully applied in evolutionary settings \citep{koutnik2010evolving, koutnik2010searching},
DCT's effect on model performance has not been clearly quantified in prior work.
We evaluate recurrent neural networks (RNNs) with DCT-encoded weight matrices on
a character level language modelling task
with various compression rates.
We also report results of the fast weight version 
in which we augment the DCT-based model by another NN which
generates the DCT-coefficients of the main network.

\begin{figure}[t]
\begin{subfigure}[b]{0.2\textwidth}
\hspace{10mm}
		\includegraphics[width=0.15\linewidth]{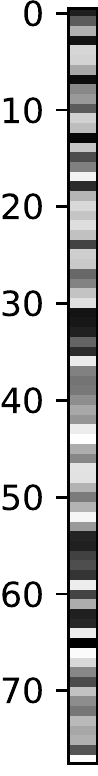}
\vspace{2mm}
\caption{DCT-coefficient vector}
\label{sfig1}
\end{subfigure}
\hspace{5mm}
\begin{subfigure}[b]{0.304\textwidth}
		\includegraphics[width=\linewidth]{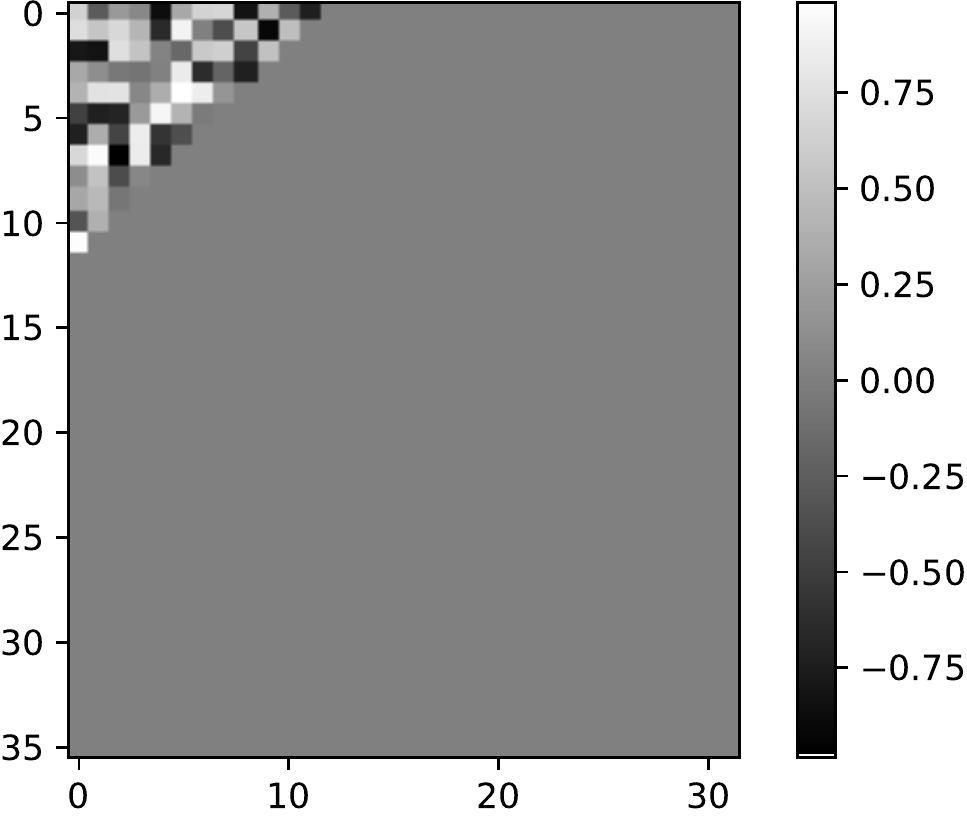}
\caption{Sparse frequency domain weight matrix}
\label{sfig2}
\end{subfigure}
\hspace{10mm}
\begin{subfigure}[b]{0.3\textwidth}
		\includegraphics[width=\linewidth]{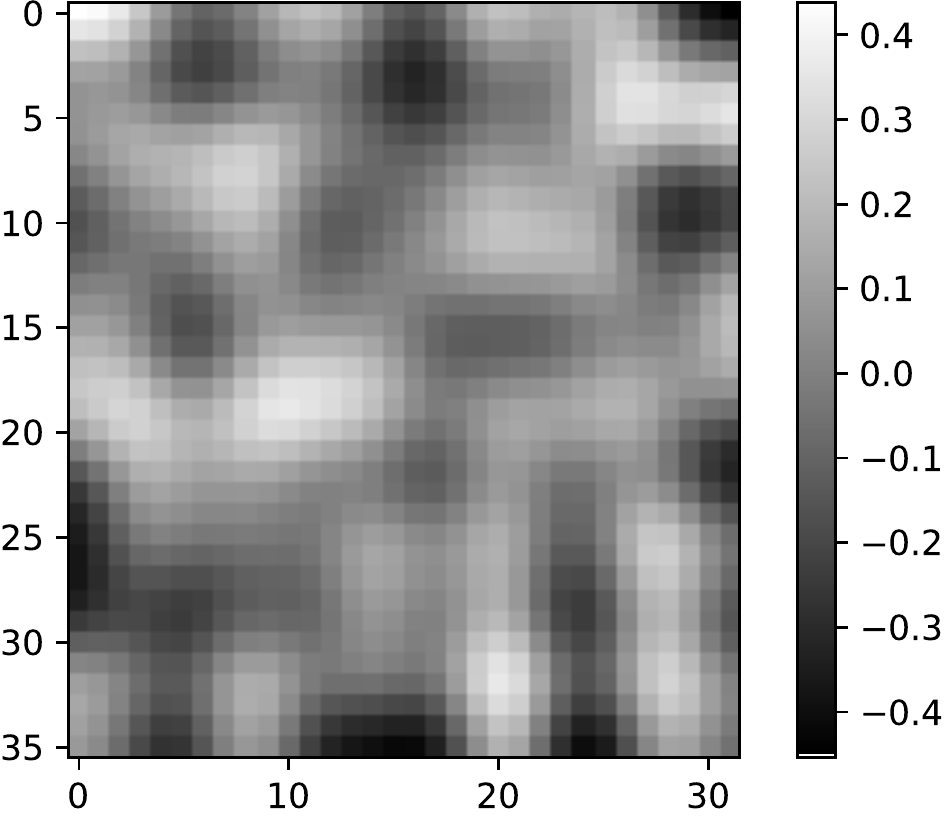}
\caption{Dense time domain weight matrix}
\label{sfig3}
\end{subfigure}
	\caption{Illustration of the decompression steps from a DCT-coefficient vector (model parameters) to an actual weight matrix.
The DCT-coefficient vector (Figure \ref{sfig1}) is transformed into sparse frequency domain weights (Figure \ref{sfig2})
by $\TopLeft$ operation (Eq.~\ref{eq:step1}).
Then, inverse DCT (Eq.~\ref{eq:step2}) transforms the frequency matrix of Figure (\ref{sfig2}) to the final dense weight matrix of Figure (\ref{sfig3}).}
	\label{fig:dct}
\end{figure}

\section{Neural Networks with DCT-Encoded Weight Matrices}
\label{sec:model}
Here we review neural networks whose weight matrices are encoded as
DCT coefficients \citep{koutnik2010evolving, koutnik2010searching, koutnik2012frequency},
and define the type of models considered in this work.

In contrast to regular neural networks whose trainable parameters are a set of weight matrices $\theta = \{\mW_1, ...,\mW_N\}$, 
trainable parameters in a DCT-based neural network are a set of DCT frequency coefficients $\mathcal{G} = \{\vg_1, ..., \vg_N\}$
which are used to generate the effective weight matrices $\{\mW^{(g_1)}, ...,\mW^{(g_N)}\}$.
In prior work \citep{koutnik2010evolving, koutnik2010searching, koutnik2012frequency, gomez2012compressed, srivastavaSG12, van2016wavelet}, $\mathcal{G}$ is referred to as \textit{genome}, and each $\vg_i$ as \textit{chromosome}.  

As a formal illustration let us consider a regular feed-forward layer (where we omit bias) which transforms 
an input vector $\vx \in \mathbb{R}^m$
into an output $\vy \in \mathbb{R}^n$ with an element-wise activation $\sigma$ and a trainable weight matrix $\mW \in \mathbb{R}^{n \times m}$ as:
\begin{align}
\vy = \sigma(\mW \vx)
\end{align}
Its DCT-based counterpart is a layer with trainable parameters $\vg \in \mathbb{R}^{c}$ with $c \ll n \times m$
which transforms $\vx \in \mathbb{R}^m$ to $\vy \in \mathbb{R}^n$ as:
\begin{align}
\mW_f &= \TopLeft_{n, m}(\vg) \label{eq:step1} \\
\mW^{(\vg)} &= \DCT^{-1}(\mW_f) \label{eq:step2} \\
\vy &= \sigma(\mW^{(\vg)} \vx)
\end{align}
where operations in Eq.~\ref{eq:step1} and \ref{eq:step2} correspond to a decompression process
which transforms a DCT-coefficient vector $\vg \in \mathbb{R}^{c}$ into a weight matrix $\mW^{(\vg)} \in \mathbb{R}^{n \times m}$.
First, the $\TopLeft_{n, m}$ operation generates a sparse matrix $\mW_f  \in \mathbb{R}^{n \times m}$
from $\vg \in \mathbb{R}^{c}$.
$\mW_f$ represents the weight matrix in the frequency domain.
Its coefficients are all zero except the $c$ highest frequency coefficients (anti-diagonals in the top left corner of the matrix)
which are filled using the coefficients of $\vg$.
This mapping is tied to the model definition, and the model is trained end-to-end under the corresponding weight pattern in the time domain.
The choice of high frequency domain (instead of low-frequency one) is empirically discussed
in Appendix \ref{app:low_vs_freq}.
The inverse of DCT, $\DCT^{-1}$ is then applied to the sparse frequency domain matrix $\mW_f$
to obtain a dense matrix $\mW^{(\vg)} \in \mathbb{R}^{n \times m}$.
Figure \ref{fig:dct} illustrates these intermediate steps of the decompression process.

We also note that both $\DCT$ and its inverse $\DCT^{-1}$ can be expressed as simple (differentiable) matrix multiplications
 using fixed DCT matrices:
\begin{align}
\DCT^{-1}(\mW_f) = D_n^{-1} \mW_f D_m^{-1}
\end{align}
where $D_n \in \mathbb{R}^{n \times n}$ and $D_m \in \mathbb{R}^{m \times m}$
are fixed orthonormalised DCT matrices.

The size $c$ of the DCT-coefficient vector is a model hyper-parameter.
With $c \ll n \times m$, where $n \times m$ is the size of the original matrix,
$\vg$ effectively offers a compressed representation of $\mW^{(\vg)}$.
In our experiments, instead of explicitly fixing $c$,
we fix a compression rate and derive the value of $c$.

In practice, we allocate separate DCT-coefficient vectors for different weight matrices (i.e.~matrices which have different roles),
e.g., in an RNN:
\begin{align}
h_t= \sigma(\mW^{(\vg_i)} x_t + \mR^{(\vg_r)} h_{t-1})
\end{align}
$\mW^{(\vg_i)} $ and $\mR^{(\vg_r)}$ are generated by parameters $\vg_i$ and $\vg_r$ respectively
via two separate transformations. Similarly, for a Long Short-Term Memory (LSTM) \citep{hochreiter1997long},
separate DCT parameters are allocated for each weight matrix and decompressed
separately.

This method has been previously investigated for RNNs
in evolutionary settings \citep{koutnik2010evolving, koutnik2010searching}
and for small feedforward NNs in supervised
settings of toy binary classification tasks \citep{fabisch2013learning}.
We revisit this method for RNNs in a language modelling
setting.
In the experimental Sec.~\ref{sec:exp},
we evaluate the performance of LSTM-RNNs with DCT-encoded weights
under different compression rates.

\begin{wrapfigure}[18]{R}{0.35\textwidth}
\begin{figure}[H]
\includegraphics[width=1\linewidth]{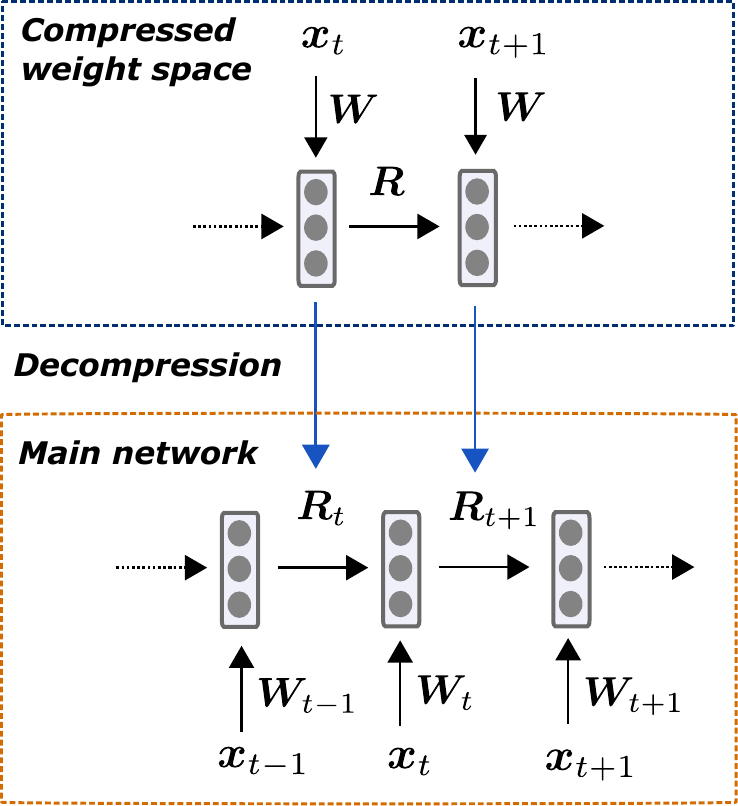}
\caption{Illustration of a network with weight matrices parametrised by an RNN in the compressed space
(weight generation for $\mW_t$ as well as super-scripts on weights are omitted for readability).}
\label{fig:fast}
\end{figure}
\end{wrapfigure}

\paragraph{Fast weight version.}
Once we have a compact representation of weight matrices,
an attractive extension is to generate the weights in the compressed space depending
on the context by parameterising
it by another neural network \citep{Schmidhuber:91fastweights, schmidhuber1992learning, schmidhuber1993reducing, GomezS05, ha2017hypernetworks, schlag2021linear}.

Here we focus on systems whose main network (fast net) is a regular RNN (Eq.~\ref{eq:fast_end})
and we parameterise its two weight matrices: input-to-hidden weights and hidden-to-hidden weights,
using two separate LSTM-RNNs (Eqs.~\ref{eq:fast_start}-\ref{eq:fast_start2}).
This system thus transforms an input vector $\vx_t \in \mathbb{R}^m$ at time step $t$ to 
a recurrent hidden state $\vh_t \in \mathbb{R}^n$ as:

\begin{align}
\vg^i_t, \vc^i_t  &= \LSTM_{\theta_i}(\vg^i_{t-1}, \vc^i_{t-1}, \vx_t) \label{eq:fast_start} \\ 
\vg^h_t, \vc^h_t  &= \LSTM_{\theta_h}(\vg^h_{t-1}, \vc^h_{t-1}, \vx_t) \label{eq:fast_start2} \\
\mW_t^{(\vg_t^i)} &= \DCT^{-1} \circ \TopLeft_{n, m} (\vg^i_t) \\
\mR_t^{(\vg_t^h)} &= \DCT^{-1} \circ \TopLeft_{n, n} (\vg^h_t) \\
\vh_t &= \sigma(\mW_t^{(\vg_t^i)} \vx_t + \mR_t^{(\vg_t^h)} \vh_{t-1}) \label{eq:fast_end}
\end{align}

where $\circ$ denotes the composition.
Figure \ref{fig:fast} illustrates the corresponding model.
This approach can be considered as an alternative to outer-product based fast weights,
with a potentially full rank weight matrix update at each time step.
Also, the slow network (e.g. $\LSTM_{\theta_i}$) can see the current states of weight matrix ($\vg^i_{t-1}$) to generate new weight states ($\vg^i_t$) in the compressed space.
In Appendix \ref{app:alternative}, we also discuss
an alternative approach where a single LSTM is used to parameterise the
two weight matrices.

\section{Experimental results}
\label{sec:exp}
We train character level LSTM-RNN language models on the standard \texttt{enwik8} dataset \citep{hutter2012human} which
contains about 90~M characters for training with a character vocabulary size of 205.
We follow the base configuration of \citet{merity2018analysis} which has three LSTM-RNN
layers with dimensions of 1840 for the two first layers and 400 for the third one,
with an embedding matrix of size 400 whose shared by the input and output layers.
In the DCT models, we encode each weight matrix in an LSTM layer by DCT as described
in Sec.~\ref{sec:model}.
We apply the same compression rate for each weight matrix.
More on experimental details can be found in Appendix \ref{app:exp}.

Table \ref{tab:training} presents the effects of different compression rates on DCT-based LSTM model performance.
We obtain degradation of about 14\% relative in bits-per-character
with a compression rate of 90\%.
This model with 4.8~M parameters has to be compared to the performance of baseline models
with smaller layer sizes (the simplest way to reduce the
model size) presented in Table \ref{tab:base}.
The performance of the baseline model with a layer size of 465 and an embedding size of 400 from Table \ref{tab:base} (also 4.8~M parameters)
is similar to the performance of  the DCT model with a compression
rate of 90\% in Table \ref{tab:training}.
Apparently a large model with compressed weights does not 
improve much over a small model
with a comparable number of parameters.
However, reducing the number of parameters by only
reducing the recurrent layer size without reducing the embedding dimension has its limits.
For example, with an LSTM layer size of 16 (which is extremely small) and
an embedding size of 400, the corresponding model size is 780~K.
Further parameter reduction requires to reduce the embedding size.
In contrast, the DCT-based model can be configured to have an arbitrarily small 
number of parameters, without requiring to modify other model hyper-parameters,
as shown in the lower part of Table \ref{tab:training}.

Table \ref{tab:fast} highlights performance of fast weight models.
The hyper-parameters are chosen such that the total number of model parameters
is comparable to those we discuss in Table \ref{tab:training} and \ref{tab:base}.
We found the performance of these models to vary a lot even with similar numbers
of parameters.
In general, we found that models with more fast variables (\# Fast Params in tables) tend to perform better.
The best model achieves performance similar to the one of baseline models.
More comparisons among the fast weight model variants can be found in Table \ref{tab:fast_ablation} in Appendix \ref{app:alternative}.

\begin{table}[t]
\RawFloats
  \begin{minipage}{.45\linewidth}
\centering
\caption{
Bits-per-character performance of \textbf{DCT-based} character level LSTM language models on \texttt{enwik8}.
Results are shown for different levels of compression rate.
The embedding size is 400 in all cases.}
\vspace{2mm}
\label{tab:training}
\begin{tabular}{lrrr} \toprule
Compression   & \# Params  & \multirow{2}{*}{Valid} &   \multirow{2}{*}{Test}  \\ 
Rate     & in M &  &    \\ \midrule
0        & 47.0    & 1.28 &  1.29       \\ 
0.5     & 23.7    & 1.33 &  1.33      \\
0.7     & 14.2    & 1.36 &   1.36      \\
0.8     &    9.5     &  1.40          &        1.40      \\
0.9     &  \textbf{4.8}   & \textbf{1.43} &   \textbf{1.44}      \\ \midrule
0.99   &  567 K   &   1.75  & 1.74 \\
0.999         &   144 K    &   2.23 & 2.20 \\ \bottomrule
\end{tabular}
\end{minipage}
\hfill
  \begin{minipage}{.49\linewidth}
\centering
\caption{
Bits-per-character performance of \textbf{baseline} character level LSTM language models on \texttt{enwik8}
with different numbers of parameters.}
\vspace{2mm}
\label{tab:base}
\begin{tabular}{rrrrr} \toprule
Embed.  & Layer & \# Params &  \multirow{2}{*}{Valid}    &   \multirow{2}{*}{Test} \\
Size  & Size & in M &   &   \\ \midrule
64        & 672 & 5.8 & 1.47 & 14.9  \\ \midrule
400 &  16 & 780~K  & 1.74 & 1.75 \\   
    & 465 & \textbf{4.8}  & \textbf{1.46} & \textbf{1.47} \\
    & 512 & \textbf{5.5}  & 1.46 & 1.47 \\ \bottomrule
\end{tabular}
\end{minipage}

\end{table}

\begin{table}[h]
\RawFloats
\vspace{-1mm}
\begin{center}
\caption{
Bits-per-character performance on \texttt{enwik8} of DCT-based models with \textbf{fast weights}.
Each weight matrix in the fast RNN is parameterised by two separate LSTMs.}
\vspace{2mm}
\label{tab:fast}
\begin{tabular}{rrrrrrrr}
\toprule
Compression & Embed. & Layer & Num & \# Fast & \# Params &  \multirow{2}{*}{Valid}  &   \multirow{2}{*}{Test} \\
Rate  & Size & Size & Layers & Params & in M &   &      \\ \midrule
0.90    &  64   & 80  & 2  & 2028 & \textbf{4.8} & 1.60 & 1.61 \\  
     & 154 & 154 & 1 & 4692 & \textbf{4.7} & 1.48 & 1.51 \\ \midrule  
0.99 &  478 & 478 & 1  & 4556 & \textbf{5.0} & \textbf{1.44} & \textbf{1.48} \\  
\bottomrule
\end{tabular}
\end{center}
\vspace{-2mm}
\end{table}


\section{Discussion and Future Work}
We revisited DCT-based indirect encodings of neural network weight matrices
in a character level language modelling task.
Our experiments seem to indicate that at least
in the case of language modelling,
large architectures with DCT-compressed weights don't improve much over
small networks with a similar number of parameters.
However, we also showed that the DCT-based encoding
allows for controlling the number of parameters in each layer
flexibly and independently of hyper-parameters in other layers. This
property may be of practical interest in some applications.

What is the best way 
of encoding 
neural network weights? This fundamental question remains unanswered, but we hope our work will help trigger a discussion of this topic.

Our fast weight variants can be extended in various ways to build
highly adaptive language models \citep{lazaridou2021pitfalls, irie18:radmm}.
In particular, the weight generator networks could be augmented by additional inputs
for error signals \citep{Schmidhuber:93selfreficann,hochreiter2001learning, rocki2016surprisal}
to facilitate the weight update generation.
Future work will evaluate our models on tasks which
explicitly require context-adaptive behaviour,
e.g., texts containing multiple domain shifts.

\section*{Acknowledgements}
We thank Sjoerd van Steenkiste and Imanol Schlag for valuable
comments and suggestions on the first version of the manuscript.
This research was partially funded by ERC Advanced grant no: 742870, project AlgoRNN,
and by Swiss National Science Foundation grant no: 200021\_192356, project NEUSYM.
We thank NVIDIA Corporation for donating several DGX
machines, and IBM for donating a Minsky machine.

\bibliography{paper}
\bibliographystyle{iclr2021_conference}
\clearpage
\appendix

\section{Experimental details}
\label{app:exp}
In this section, we provide experimental details.
As a baseline architecture to evaluate DCT-parameterised NNs (Table \ref{tab:training}),
we follow the base configuration of \citet{merity2018analysis} which has three LSTM-RNN
layers with 1840 nodes for the two first layers and 400 for the third one,
with an embedding layer of size 400 whose parameters are shared for input and output embeddings.
For all small models with less than 6~M parameters in all tables, dropout is completely disabled to avoid an extra factor for comparison.
For larger models in Table \ref{tab:training}, the dropout configuration from 
\citet{merity2018analysis} is applied: 0.1 to the feed-forward hidden layers
between the LSTM layers, 0.2 to the recurrent weights, and 0.4 to the output of the last LSTM layer.
Otherwise all models in all tables are trained with the same training configuration:
we use Adam optimiser with a learning rate of 0.001, with a batch size of 128,
and a backpropagation span of 200 characters.

In the DCT models, we encode each weight matrix (separately for each gate) of an LSTM layer by DCT as described
in Sec.~\ref{sec:model}.
We apply the same compression rate for all weight matrices.
To initialise the model parameters which are DCT-coefficient vectors, we first generate and initialise a weight matrix in time domain,
and apply the DCT compression to obtain the corresponding initialisation,
which we found to work well in practice.

\paragraph{Implementation notes.}
Like any fast weight variants which generate the entire weight matrix at each time step,
implementation of DCT based fast weight RNNs requires a custom operation for backpropagation.
In fact, a naive implementation would store weights for each time step for the backward pass,
which can quickly result in prohibitive space requirement.
Instead, in a custom implementation we only store the compressed weights for each time step
during the forward pass, and the actual weights are recomputed via decompression
for each backward step.
An obvious drawback is that the decompression operations need to be called at each backward step,
which can be slow.
The corresponding code is available at {\small \url{https://github.com/kazuki-irie/dct-fast-weights}} .

\section{Ablation studies and extra experimental results}
Here we present extra experimental results which could not be included in the main text
because of the space limitation.

\subsection{Choice of high vs low frequency sparsity patterns}
\label{app:low_vs_freq}
The choice of high frequency sparsity pattern in the frequency matrix is arbitrary for our language models,
because the textual data does not contain any a priori frequency property, and
the input character embeddings are jointly learned with the rest of model parameters,
given the pre-determined architectural choice of high or low-frequency coefficients in the model.
Our choice is thus only supported by the empirical results, presented in Table \ref{tab:low_vs_freq},
which show that the DCT models with high-frequency non-zero coefficients tend to outperform those using low-frequency coefficients. 

\begin{table}[h]
\RawFloats
\begin{center}
\caption{Comparison of DCT-based language models with high vs.~low frequency coefficients. Language model bits-per-character performance on \texttt{enwik8}.}
\vspace{2mm}
\label{tab:low_vs_freq}
\begin{tabular}{rrrrr} \toprule
Compression & Coefficient  & Number  & \multirow{2}{*}{Valid} &   \multirow{2}{*}{Test}  \\ 
Rate  & Frequency  & Params. &  &    \\ \midrule
0.90   & Low  &       4.8~M              & 1.49 &   1.50      \\ 
        & High  &  & \textbf{1.43} &   \textbf{1.44}      \\ \midrule
0.99 & Low  &  567~K  &   1.78  & 1.77 \\  
        & High &  &  \textbf{1.74}  & \textbf{1.74} \\ \bottomrule
\end{tabular}
\end{center}
\end{table}

\subsection{Alternative fast weight parameterisations}
\label{app:alternative}
In Eqs.~\ref{eq:fast_start}-\ref{eq:fast_end}, we used two separate slow LSTM RNNs to parameterise DCT coefficients of
the two weight matrices of the fast RNN.
An alternaitve approach is to parameterise the two DCT vectors as the output of a single slow LSTM,
such that it is aware of the states of all weights of the fast RNN to generate new weights.
The corresponding equations are as follows:
\begin{align}
[\vg^i_t, \vg^h_{t-1}], \vc^i_t &= \LSTM([\vg^i_{t-1}, \vg^h_{t-1}], \vc^i_{t-1}, \vx_t, ) \\
\mW_t^{(\vg_t^i)} &= \DCT^{-1} \circ \TopLeft_{n, m} (\vg^i_t) \\
\mR_t^{(\vg_t^h)} &= \DCT^{-1} \circ \TopLeft_{n, n} (\vg^h_t) \\
\vh_t &= \sigma(\mW_t^{(\vg_t^i)} \vx_t + \mR_t^{(\vg_t^h)} \vh_{t-1})
\end{align}

However, as shown in Table \ref{tab:fast_ablation}, 
we experimentally found that the variant with two separate LSTMs
(which we denote as \textit{Twin} approach) perform better
than the variant using only one LSTM (denoted as \textit{Single}).
Thus, in Table \ref{tab:fast}, we only reported results for \textit{twin}
model variants.

\begin{table}[h]
\RawFloats
\begin{center}
\caption{Comparison of DCT-based \textbf{fast weight} models with single vs.~twin LSTM parameterisation
of fast RNN. Language model bits-per-character performance on \texttt{enwik8}.}
\vspace{2mm}
\label{tab:fast_ablation}
\begin{tabular}{rrrrrrrrr}
\toprule
Compression & \multirow{2}{*}{Slow LSTM} & Embed. & Hidden & Num & \# Fast & \# Params &  \multirow{2}{*}{Valid}  &   \multirow{2}{*}{Test} \\
Rate  &  & Size & Size & Layers & Params & in M &   &      \\ \midrule
0.90 & Single & 64 & 64 & 3 & 812 & 4.6 & 1.64 & 1.65 \\  
    & & 64 & 80  & 2  & 1126 &4.8 & 1.66 & 1.66 \\  
    &  & 130 & 130 & 1  & 3306 & 4.5 & \textbf{1.64} & \textbf{1.65} \\ \cmidrule(r){2-9} 
    & Twin &  64   & 80  & 2  & 2028 & 4.8 & 1.60 & 1.61 \\  
    &  & 154 & 154 & 1 & 4692 & 4.7 & \textbf{1.48} & \textbf{1.51} \\  \midrule 
0.99 & Single & 64 & 306 & 2  & 1319 &  5.5 & 1.59 & 1.61 \\  
        & Twin &  478 & 478 & 1 & 4556 & 5.0 & \textbf{1.44} & \textbf{1.48} \\  
\bottomrule
\end{tabular}
\end{center}
\end{table}

\end{document}